\documentclass[runningheads]{llncs}

\usepackage{preamble}

\begin{document}

\title{Reproducible Multimodal Affordance Prediction}

\titlerunning{Reproducible Multimodal Affordance Prediction}

\author{Tommaso Apicella\inst{1}\orcidlink{0000-0001-9001-5641} \and
Alessio Xompero\inst{2}\orcidlink{0000-0002-8227-8529} \and
Andrea Cavallaro\inst{3}\orcidlink{0000-0001-5086-7858}}

\authorrunning{T.~Apicella et al.}

\institute{Istituto Italiano di Tecnologia, Italy 
\and Independent Researcher 
\and EPFL, Switzerland\\
\email{t.apicella.cs@gmail.com}, \email{alessio.xompero@gmail.com}, \email{andrea.cavallaro@epfl.ch}}

\maketitle

\begingroup
\renewcommand\thefootnote{}
\footnotetext{\scriptsize 
A. Xompero did part of the work when he was affiliated with Queen Mary University of London.}
\endgroup

\begin{abstract}
Affordance prediction is the identification of potential actions an agent can perform on a target object from multimodal inputs. Affordance prediction methods are difficult to evaluate and compare due to heterogeneous problem formulations, inconsistent dataset annotations, incomplete reporting of experimental protocols, and limited information about deployment conditions. These limitations challenge fair benchmarking and performance comparison. To promote transparency, we propose the Affordance Sheet, a documentation detailing task formulation with its input modalities, model architectures and training information, datasets, and experimental protocols. Affordance Sheets enable reproducible benchmarking and reliable evaluation of affordance models for real-world scenarios, including generalisation to novel conditions and human safety. 

\keywords{Affordance \and Benchmarking \and Transparency}
\end{abstract}

\section{Introduction}
\label{sec:introduction}

\setcounter{footnote}{0}

Affordances describe the potential actions that an agent can perform on objects in the environment~\cite{gibson1966senses}, inferred from multimodal sensory observations such as visual, depth, acoustic, and tactile data~\cite{jamone2016affordances}. Understanding affordances enables the agent to accomplish a task, selecting which objects in the environment to interact with, what actions to perform, and how to execute them. 
This reasoning within and across modalities, beyond simply perceiving scene and objects, supports applications such as human-robot collaboration~\cite{sanchez2020benchmark} and wearable robotics~\cite{castro2022continuous}, characterised by unstructured environments and occlusions.

\begin{figure}[t!]
    \centering
    \includegraphics[width=0.9\linewidth]{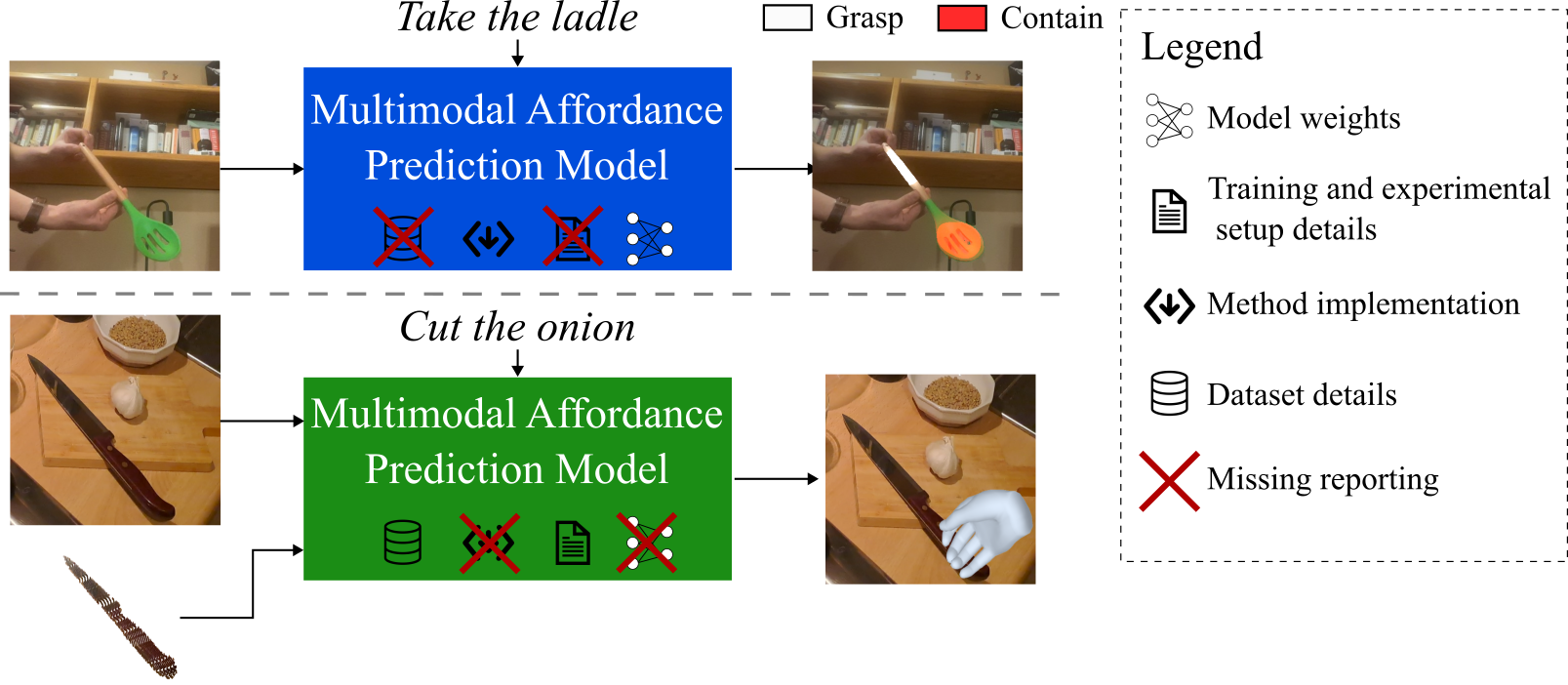}
    \caption{Illustrative example of affordance prediction from multimodal inputs (e.g. images, language, point clouds) formulated differently, such as segmentation (top) or hand pose estimation (bottom), and lacking transparent details such as code implementation, model weights, or training and experimental setup. Different reporting makes models not reproducible or comparable, and prevents their deployment in diverse real-world environments and in human-centred scenarios (e.g. human-robot interaction).}
    \label{fig:formulations}
\end{figure}

The generic definition of affordance has resulted in different formulations for the prediction of affordances, capturing only a partial aspect of the overall problem (see Fig.~\ref{fig:formulations}). Affordance classification~\cite{nagarajan2020ego} focuses on identifying plausible actions. Affordance segmentation~\cite{do2018affordancenet} aims to localize relevant interaction regions on objects. Hand pose estimation~\cite{lundell2021multi} predicts hand configurations that enable the actions. This fragmentation is worsened by the use of datasets focused on specific sensors, with partially overlapped annotation, limited code availability, inconsistent training setups, and partially reported evaluation protocols. Moreover, the design of methods and learning-based models is mostly done in controlled settings with objects placed fully visible in laboratory-like scenes (e.g. a tabletop) and considering a limited number of object types~\cite{myers2015affordance, jiang2022a4t, apicella2023affordance, nguyen2017object,guo2023handal, khalifa2023large}. Multimodal affordance prediction models should be designed and evaluated to assess their \textit{generalization} to diverse objects and real-world scenarios (including interactions with people manipulating the target object), their \textit{robustness} to occlusions from other objects or from the hand of a person holding the object, and \textit{safety} for the person~\cite{pang2021towards}.
All the above factors result in misleading and incomplete insights, slowing progress in the field and limiting the benchmarking of models for reliable real-world deployment.

\begin{figure}[t!]
    \centering
    \scriptsize
    \setlength\tabcolsep{0.1pt}
    \begin{tabular}{ccc}
    \includegraphics[height=0.245\linewidth, width=0.33\linewidth]{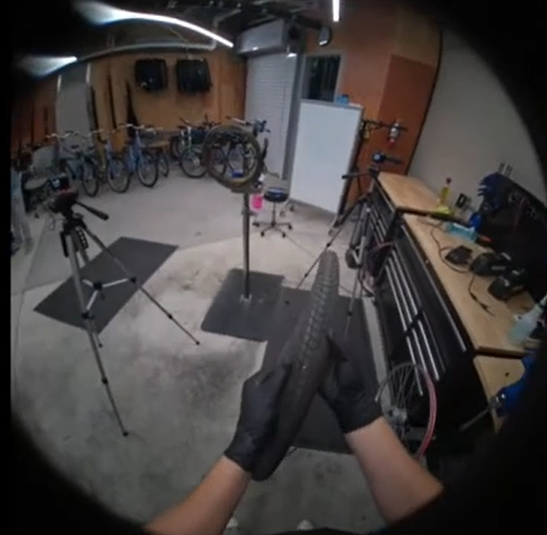} &
    \includegraphics[height=0.245\linewidth]{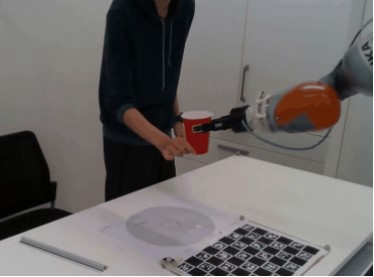} &
    \includegraphics[height=0.245\linewidth]{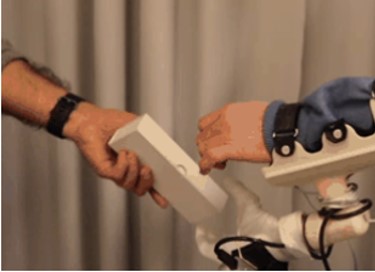} \\
    \end{tabular}
     \caption{Examples of applications benefiting from affordance prediction: immersive tutorial~\cite{grauman2024ego} (left), human-robot collaboration~\cite{sanchez2020benchmark} (centre), and human-to-human collaboration with wearable robotics~\cite{castro2022continuous} (right).
     }
    \label{fig:applications}
\end{figure}

\textit{Reproducibility}, defined as obtaining the same results given the same conditions (i.e. data, training and testing setups, and trained model)~\cite{pineau2021improving}, is a central concern in machine learning~\cite{pineau2021improving} with works highlighting inconsistencies in experimental practices and proposing standardized reporting tools~\cite{gebru2021datasheets, mitchell2019model}. 
Model cards~\cite{mitchell2019model} and datasheets~\cite{gebru2021datasheets} aim to improve transparency when releasing machine learning models and datasets, respectively. Model cards~\cite{mitchell2019model} help the identification of model limitations by describing the method, the experimental setup, and the applications or conditions leading to underperformance. Datasheets for datasets~\cite{gebru2021datasheets} facilitate the description of data limitations by detailing dataset collection and annotation procedures, intended use, and potential biases. These reporting practices have been used to document multimodal models such as Vision-Language-Action Models (VLAs)~\cite{lerobot_smolvla_base, lerobot_pi05_base} or World Action Models~\cite{microsoft_wham, dreamzero_droid} and datasets, providing details and intended use about methods and data.
However, these data- and model-centric tools do not capture the task-specific ambiguities that characterize affordance prediction, such as the diversity of problem formulations, the definition and semantics of affordances, and the characteristics of agent–object interactions. As a result, critical aspects required for fair comparison, such as how affordances are defined and how predictions translate into actionable behaviours, are not specified. Additionally, model cards and datasheet do not explicitly account for the validation of methods through in-the-wild conditions or through deployment (e.g. on a real or simulated robot) to assess the generalisation and robustness of the performance.
Reproducibility in affordance prediction is critical because learned models can directly perform physical interactions with the environment. Non-reproducible results compromise robustness to occlusions, safety, and generalization to diverse real-world scenarios. Ensuring reproducibility therefore supports both the credibility of research findings and the development of reliable systems capable of consistent and safe interaction with humans in real-world environments (see Fig.~\ref{fig:applications}). 

In this work, we surface reproducibility issues of current affordance works, which lead to inconsistent comparisons across methods. To address these challenges, we introduce the Affordance Sheet, an Open Science document inspired by Model Cards~\cite{mitchell2019model}, detailing the tackled task, the dataset, the method, and the experimental setup and validation. Additionally, we provide and discuss four Affordance Sheet of state-of-the-art models for affordance prediction. Affordance Sheet enables to catalogue solutions designed to work in unconstrained settings, compare characteristics and validations across methods, and improve reproducibility and promote transparency\footnote{Project webpage at \url{https://apicis.github.io/aff-sheet}.}. The tool is generic and can be extended to methods such as VLAs and affordance foundation models, and to tasks beyond affordance prediction.

\section{Reproducibility Challenges}

\begin{figure*}[t!]
    \centering
    \scriptsize
    \setlength\tabcolsep{0.1pt}
    \includegraphics[width=\linewidth]{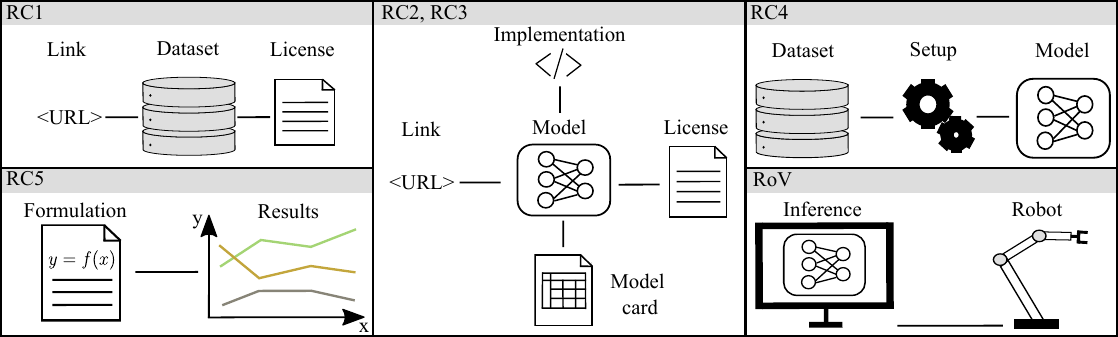}
    \caption{Visualization of the main challenges tackled by Affordance Sheet sections. KEYS -- RC:~reproducibility challenge, RoV:~robot validation}
    \label{fig:research_challenges}
\end{figure*}

Reproducibility challenges (RCs) of affordance prediction (see Fig.~\ref{fig:research_challenges}) include: 
\begin{enumerate}
    \item data availability for benchmarking (RC1);
    \item availability of a method's implementation (RC2);
    \item availability of trained models (RC3);
    \item details of experimental setups (RC4); and
    \item details of performance measures for evaluation (RC5). 
\end{enumerate} 

\subsection{Datasets and Benchmarks}

Table~\ref{tab:affordance_datasets} reports the main characteristics of the available datasets and benchmarks in  affordance prediction. None of the datasets is collected for benchmarking methods under different in-the-wild conditions (RC1), such as illumination, clutter, or hand-occlusions. 
Each affordance dataset is collected for a specific formulation, and not re-used across different affordance tasks, limiting fair comparison and preventing comprehensive benchmarking. One of the main obstacles to obtain large-scale data collections to assess models performance is the annotation. Precise annotations are manual and time-consuming, limiting the size of most datasets to less than 50,000 images and less than 10 affordance categories. For this reason, the largest datasets, such as HANDAL~\cite{guo2023handal} or CHOC-AFF~\cite{apicella2023affordance}, either used synthetic data to scale the annotation or adapted methods to provide weakly or self-supervised annotation.

\begin{table}[t!]
    \scriptsize
    \setlength\tabcolsep{2pt}
    \centering
    \caption{Characteristics of datasets for affordance prediction grouped by formulations.}
    \begin{tabular}{c c ccc r c c c c c}
         \toprule
         \textbf{Formulation} & \textbf{Dataset} & \textbf{VIS} & \textbf{LAN} & \textbf{PCL} & \textbf{\# Images} & \textbf{OBJ} & \textbf{AFF} & \textbf{Real} & \textbf{Tran.} & \textbf{HOc} \\
         \midrule
         \multirow{4}{*}{\textit{AFFC}}
         & Pieropan et al.~\cite{pieropan2013functional} & \bbox & \wbox & \wbox  & $\sim$40,000 & 4 & 4 & \bbox & \wbox & \wbox \\ 
         & Zheng et al.~\cite{zheng2018high} & \bbox & \wbox & \wbox & 740 & 8 & 3 & \bbox & \wbox & \wbox \\ 
         & Sun et al.~\cite{sun2010learning} & \bbox & \wbox & \wbox & ~1400 & 7 & 6 & \bbox & \wbox & \wbox \\ 
         & Kjellström et al.~\cite{Kjellstrom2011visual} & \bbox & \wbox & \wbox & ~11,500 & 6 & 3 & \bbox & \wbox & \raisebox{1pt}{\scalebox{0.5}{\LEFTcircle}} \\
         \midrule
         \multirow{14}{*}{\textit{AFFDS}}
         & AFF-Synth~\cite{christensen2022learning} & \bbox & \wbox & \wbox & 30,245 & 21 & 7 & \wbox & \wbox & \wbox \\
         & UMD-Synth~\cite{chu2019learning} & \bbox & \wbox & \wbox & 37,200 & 17 & 7 & \wbox & \wbox & \wbox \\
         & Multi-View~\cite{khalifa2023large} & \bbox & \wbox & \wbox & 47,210 & 37 & 15 & \bbox & \wbox & \wbox \\
         & HANDAL~\cite{guo2023handal} & \bbox & \wbox & \wbox & 308,000 & 17 & 1 & \bbox & \wbox & \raisebox{1pt}{\scalebox{0.5}{\LEFTcircle}}  \\
         & TRANS-AFF~\cite{jiang2022a4t} & \bbox & \wbox & \wbox & 1,346 & 3 & 3 & \bbox & \bbox & \wbox \\
         & UMD~\cite{myers2015affordance} & \bbox & \wbox & \wbox & 28,843 & 17 & 7 & \bbox & \wbox & \wbox \\ 
         & IIT-AFF~\cite{nguyen2017object} & \bbox & \wbox & \wbox & 8,835 & 10 & 9 & \bbox & \raisebox{1pt}{\scalebox{0.5}{\LEFTcircle}} & \raisebox{1pt}{\scalebox{0.5}{\LEFTcircle}} \\
         & CAD120-AFF~\cite{sawatzky2017weakly} & \bbox & \wbox & \wbox & 3,090 & 11 & 6 & \bbox & \wbox & \raisebox{1pt}{\scalebox{0.5}{\LEFTcircle}} \\ 
         & FPHA-AFF~\cite{hussain2020fpha} & \bbox & \wbox & \wbox & 4,300 & 14 & 8 & \bbox & \raisebox{1pt}{\scalebox{0.5}{\LEFTcircle}} & \bbox \\
         & EPIC-AFF~\cite{mur2023multi} & \bbox & \wbox & \bbox & 38,876 & 304 & 43 & \bbox & \raisebox{1pt}{\scalebox{0.5}{\LEFTcircle}} & \bbox \\
         & CHOC-AFF~\cite{apicella2023affordance} & \bbox & \wbox & \wbox & 138,240 & 3 & 3 & \raisebox{1pt}{\scalebox{0.5}{\LEFTcircle}} & \raisebox{1pt}{\scalebox{0.5}{\LEFTcircle}} & \raisebox{1pt}{\scalebox{0.5}{\LEFTcircle}} \\
         & 3DAffordanceNet~\cite{deng20213d} & \wbox & \wbox & \bbox & - & 23 & 18 & \wbox & \wbox & \wbox \\
         & LASO~\cite{li2024laso} & \wbox & \bbox & \bbox & - & 23 & 17 & \wbox & \wbox & \wbox \\
         & SceneFun3D~\cite{delitzas2024scenefun3d} & \bbox & \bbox & \bbox & - & - & 9 & \bbox & \wbox & \wbox \\
         \midrule
         \multirow{4}{*}{\textit{AFFG}}
         & OPRA~\cite{demo2vec2018cvpr} & \bbox & \wbox & \wbox & - & - & 7 & \bbox & \wbox & \wbox \\
         & AGD20K~\cite{luo2022learning} & \bbox & \wbox & \wbox & 23,816 & 47 & 36 & \bbox & \raisebox{1pt}{\scalebox{0.5}{\LEFTcircle}} & $\raisebox{1pt}{\scalebox{0.5}{\LEFTcircle}}$ \\
         & PIAD~\cite{yang2023grounding} & \bbox & \wbox & \bbox & - & 23 & 17 & \bbox & \wbox & \wbox  \\
         & MIPA~\cite{gao2025learning} & \bbox & \wbox & \bbox & - & 23 & 17 & \bbox & \wbox & \wbox \\
         \midrule
         \multirow{4}{*}{\textit{GDET}}
         & Cornell grasping~\cite{jiang2011efficient} & \bbox & \wbox & \wbox & 1,035 & - & 1 & \bbox & \wbox & \wbox \\ 
         & GraspSeg~\cite{asif2018graspnet} & \bbox & \wbox & \wbox & 33,188 & 15 & 1 & \bbox & \wbox & \wbox \\
         & Jacquard~\cite{depierre2018jacquard} & \bbox & \wbox & \wbox & 54,485 & - & 1 & \wbox & \wbox & \wbox \\
         & OCID~\cite{ainetter2021end, suchi2019easylabel} & \bbox & \wbox & \wbox & - & - & 1 & \bbox & \wbox & \wbox \\
         \midrule
         \multirow{6}{*}{\textit{HPE}}
         & EPIC-Kitchens~\cite{damen2018scaling} & \bbox & \bbox & \wbox & - & - & 1 & \bbox & \wbox & \wbox \\
         & YCB-Affordance~\cite{corona2020ganhand} & \bbox & \wbox & \wbox & 133,936 & 58 & 1 & \bbox & $\raisebox{1pt}{\scalebox{0.5}{\LEFTcircle}}$ & \wbox \\ 
         & HO3Pairs~\cite{ye2023affordance} & \bbox & \wbox & \wbox & - & - & 1 & \bbox & \wbox & $\raisebox{1pt}{\scalebox{0.5}{\LEFTcircle}}$ \\
         & GraspNet-1Billion~\cite{fang2020graspnet} & \bbox & \wbox & \bbox & - & 88 & 1 & \bbox & \wbox & \wbox \\
         & DexGraspNet~\cite{wang2022dexgraspnet} & \wbox & \wbox & \bbox & - & 133 & 1 & \wbox & \wbox & \wbox \\
         & 1M-HUGs~\cite{wu2026human} & \bbox & \wbox & \bbox & - & - & 1 & \bbox & \wbox & \wbox \\
         \bottomrule \addlinespace[\belowrulesep]
         \multicolumn{11}{l}{\parbox{\linewidth}{\scriptsize{KEYS -- VIS:~vision with RGB(-D), LAN:~language, PCL:~point cloud, \# Images:~number of images, OBJ:~number of object categories, AFF:~number of affordance categories, Tran.:~transparency, 3PV:~third person view, HOc:~hand-occlusion; 
         AFFC:~affordance classification; AFFG:~affordance grounding; HPE:~hand pose estimation; GDET:~grasping detection; AFFDS:~affordance detection and segmentation; \bbox: considered, \wbox: not considered, \raisebox{1pt}{\scalebox{0.4}{\LEFTcircle}}: partly considered.}}}
     \end{tabular}
     \label{tab:affordance_datasets}
 \end{table}

Most of the affordance datasets~\cite{ye2023affordance,myers2015affordance,asif2018graspnet,depierre2018jacquard,chu2019learning,khalifa2023large,jiang2011efficient,lakani2019towards} are acquired in controlled laboratory environments. A single \textit{unoccluded} object is placed on a planar surface (e.g. a tabletop), with fixed camera viewpoints and illumination, while varying only the object category or instance. 
For instance, both UMD~\cite{myers2015affordance} and Multi-View~\cite{khalifa2023large} include multiple object categories collected under pre-defined acquisition settings, such as constant lighting and rotating platforms. These controlled conditions simplify data collection and evaluation but do not adequately capture the variability encountered during real-world deployment, limiting the generalization of trained models to cluttered scenes, diverse backgrounds, or illumination variations.
Other datasets~\cite{nguyen2017object,guo2023handal,corona2020ganhand,sawatzky2017weakly,luo2022learning,apicella2023affordance} included images of objects with occlusions caused by surrounding objects or human hands interacting with the scene to increase realism. Such conditions substantially increase the difficulty of affordance prediction while better reflecting practical robotic applications. Hand-induced occlusions are common in human-robot collaboration scenarios, where inaccurate affordance estimation may result in unsafe or unintended interactions, posing risks to human operators~\cite{pang2024stereo,apicella2023affordance}.

Most of previous works~\cite{nguyen2016detecting,do2018affordancenet, gu2021visual, zhao2020object, zhang2022multi, yin2022object} trained models on the training split of a specific dataset and then compared the models performance on the testing split of one or more datasets.
Cross-dataset evaluations are mostly avoided due to partial overlapping of affordance classes or of object categories, across the selected datasets~\cite{apicella2024segmenting}. Datasets such as UMD~\cite{myers2015affordance}, IIT-AFF~\cite{nguyen2017object}, and Multi-View~\cite{khalifa2023large} share some of the object and affordance classes but labelled with different conventions, making the comparison of models trained on different datasets difficult. As a consequence, researchers train multiple versions of the same model, adapted to the classes of a specific dataset.
Additional documentation, such as metadata, help researchers train or evaluate methods only on common categories by re-ordering them. 
Moreover, relying only on a single benchmark can lead to limited and not generalisable considerations on model rankings. For example, images in UMD and Multi-View are collected in a laboratory environment with static conditions, e.g. a fixed camera oriented towards a table where an object is placed (camera-object distance is almost always the same)~\cite{myers2015affordance, khalifa2023large}. However, in real scenarios the camera might be closer or farther from objects compared to the training setting, hence the performance on the benchmark might not reflect in-the-wild performance.

\subsection{Methods and Experimental Setups}
The lack of publicly available implementation of methods (RC2)~\cite{gu2021visual,zhao2020object,zhang2022multi,yin2022object}, the lack of publicly available trained models (RC3)~\cite{nguyen2016detecting,gu2021visual,zhao2020object,zhang2022multi,yin2022object}, and the lack of details of experimental setups (RC4)~\cite{nguyen2016detecting,do2018affordancenet,gu2021visual,zhao2020object,zhang2022multi,yin2022object}   
can challenge researchers in reproducing previous works for comparative evaluations. 
The release of the model trained weights and the implementation of the method and inference pipeline is a crucial aspect for reproducibility, especially for deep-learning based models, allowing other researchers to test models on their own data without re-training. The availability of model implementation and weights is important when researchers need a comparison, as re-training the model can be too time- and resource-consuming. In case a new dataset is proposed and a previous method needs re-training, only the method implementation is sufficient. The re-implementation of methods and setup is time-consuming and prone to errors, and not always leads to the expected outcome (i.e. results are not replicable or findings are not reproducible). To avoid this issue and save time, researchers report results from previous works~\cite{gu2021visual,zhao2020object,zhang2022multi,yin2022object}, resulting in misleading findings and conclusions when different  experimental conditions are used.

Using the same \textit{experimental setup} to train and test affordance models allows a fair comparison and validation of new technical contributions. When releasing the training and testing code is not possible, reporting all details to reproduce a setup becomes fundamental.
The experimental setup details include training hyper-parameter values, chosen data splits, image pre-processing (normalisation and cropping procedures), and post-processing. The lack of experimental setup details causes methods for affordance detection and segmentation to be not reproducible~\cite{do2018affordancenet,nguyen2016detecting,zhang2022multi,yin2022object,gu2021visual,zhao2020object}. 
For example, AffordanceNet and BPN did not include image resize during training and testing phases~\cite{do2018affordancenet, yin2022object}, whereas DRNAtt, RANet, and GSE did not include these details for the testing phase. Gu et al.'s work~\cite{gu2021visual} omitted the parameters of the optimizers used during training. 
Apicella et al.'s work~\cite{apicella2024segmenting} showed that the lack of details in the experimental setup led to unfair comparisons. 

\begin{table}[t!]
    \centering
    \scriptsize
    \setlength\tabcolsep{2pt}
    \caption{Comparison of training/testing setups used by different methods for affordance detection and segmentation on the UMD dataset~\cite{myers2015affordance}. 
    Due to the setup inconsistencies, direct comparison among models performance is unfair.
    }
    \begin{tabular}{c c cccc cc}
    \toprule
    \textbf{Training setup} & \textbf{Resolution} & \multicolumn{4}{c}{\textbf{Data augmentation}} & \multicolumn{2}{c}{\textbf{Image resize}} 
    \\
    \cmidrule(lr){3-6}\cmidrule(lr){7-8}
    & & FLIP & SCALE & ROT & JIT & Train. & Test.\\
    \midrule
    AffordanceNet~\cite{do2018affordancenet} & $1000 \times 600$ & $\circ$ & $\circ$ & $\circ$ & $\circ$ & unknown & unknown
    \\ 
    CNN~\cite{nguyen2016detecting} & $320 \times 240$ & $\circ$ & $\circ$ & $\circ$ & $\circ$ & centre-crop & sliding window
    \\    
    DRNAtt~\cite{gu2021visual} & $320 \times 240$ & $\circ$ & $\circ$ & $\circ$ & $\circ$ & centre-crop & unknown
    \\ 
    RANet~\cite{zhao2020object} & $224 \times 224$ & $\circ$ & $\circ$ & $\circ$ & $\circ$ & centre-crop & unknown 
    \\
    GSE~\cite{zhang2022multi} & $400 \times 400$ & $\bullet$ & $\bullet$ & $\circ$ & $\circ$ & crop & unknown 
    \\    
    BPN~\cite{yin2022object} & $1000 \times 600$  & $\bullet$ & $\bullet$ & $\bullet$ & $\bullet$ & unknown & unknown
    \\
    \bottomrule
    \addlinespace[\belowrulesep]
    \multicolumn{8}{l}{\parbox{0.9\linewidth}{\scriptsize{KEYS -- $\bullet$:~considered, $\circ$:~not considered, FLIP:~flipping, SCALE:~scaling, ROT:~rotating, JIT:~colour jittering, Train.:~training set, Test.:~testing set}}}\\
    \end{tabular}
    \label{tab:umdtrainingsetup_pre}
\end{table} 

Table~\ref{tab:umdtrainingsetup_pre} reports the training and testing setups of affordance detection and segmentation methods on the UMD dataset.  
Despite being trained and tested on the same dataset, models' performance is not directly comparable due to inconsistencies in the setups such as the image resize procedure and augmentation procedure during training. 
Inconsistencies can also be present in previous methods adapted into a baseline to compare with. For example, AffordanceDiffusion~\cite{ye2023affordance} is compared with the coarse hand prediction of GanHand~\cite{corona2020ganhand} due to the missing annotation of the object pose in the dataset. 
However, since a part of the architecture and of the training procedure is missing, the result is only a proxy to the (unknown) performance of GanHand.

The redefinition of the affordance problem can also result in experimental validations that ignore datasets and benchmarks of partially overlapping formulations. For example, works on affordance grounding~\cite{luo2022learning} do not compare the performance of proposed methods with that of affordance segmentation methods~\cite{nguyen2016detecting,zhang2022multi,gu2021visual,zhao2020object}, despite the similar problem formulation~\cite{luo2022learning,nguyen2016detecting}. Methods for affordance segmentation output a binary mask for each action in a predefined set of classes, whereas methods for affordance grounding output a confidence map describing where an action known a priori can take place in the image. Despite these differences, comparing methods for both affordance grounding and affordance detection and segmentation can explain if using action as input (affordance grounding) to a model provides any advantage.

\subsection{Performance Measures}

The performance of different methods for each affordance formulation~\cite{nagarajan2020ego,do2018affordancenet,lundell2021multi} was evaluated using scores or metrics to quantify the discrepancy between predictions and annotations. Describing a performance measure help other researchers understand if the experiment validates their claim or if a different measure should be chosen. Providing the mathematical formulation of the measures helps disambiguate different implementations of the same score, especially when a public evaluation toolkit is not used or referred to. For example,  mean \textit{IoU} can be the average of all the \textit{IoU}s between prediction and annotation, or the \textit{IoU} considering the full set of predictions and annotations. 
Previous works evaluated a few methods with different performance measures or datasets, making comparison and ranking not possible. For example, the performance of STRAP~\cite{cui2023strap} was measured using \textit{IoU} instead of $F^w_\beta$, as most of other methods do~\cite{do2018affordancenet,nguyen2016detecting,gu2021visual,zhao2020object,zhang2022multi,yin2022object}.

\section{Affordance Sheets}

To support reproducibility in affordance prediction, we propose the Affordance Sheet, an organised collection of good practices favouring fair comparisons. Each section of the Affordance Sheet (see Table~\ref{tab:affordance_sheet}) addresses a specific reproducibility challenge providing an informative tool to catalogue and compare methods.

The first section identifies which problems the affordance model tackles, the inputs used by the model, and the presence of the human in the scenario. This section helps researchers understand what are the competing methods and assess their performance of solutions under the same inputs and conditions.
When proposing a new problem partially overlapping with another one, previous models can be used or adapted to validate the method. For example, selecting the channel of an affordance segmentation output based on the action considered by the affordance grounding method enables the comparison between methods for affordance segmentation and methods for affordance grounding. To compare the grounding and segmentation outputs, the grounding confidence map can be converted to a binary mask via thresholding. Alternatively, the segmentation map can be converted to a confidence map by using Gaussian blur~\cite{luo2022learning}. In the sheet, we unify affordance detection and segmentation with affordance grounding under functional segmentation; we consider affordance classification as functional classification, and grasping detection as a case of hand pose estimation. 

\begin{table}[t!]
    \centering
    \setlength\tabcolsep{2pt}
    \scriptsize
    \caption{Affordance Sheet, inspired by Model Cards~\cite{mitchell2019model}, to favour transparency and reproducibility of visual affordance prediction methods.}
    \begin{tabular}{|c c c c c c c c|}
    \hline

    \rowcolor{gray!50}
    \multicolumn{8}{|c|}{\textbf{Model name - access date (dd/mm/yyyy)}} \\
    \hline

    \rowcolor{gray!25}
    \multicolumn{8}{|l|}{\textbf{Formulation}} \\
    \multicolumn{1}{|l}{\textit{Affordance task:}} 
    & FUNC & FUNS & HPE & HIS & & & \\
    & \wbox & \wbox & \wbox & \wbox & & & \\
    \multicolumn{1}{|l}{\textit{Modality:}} & Image & Video & Language & Point cloud & Depth & Audio & Tactile \\
     & \wbox & \wbox & \wbox & \wbox & \wbox & \wbox & \wbox \\
    \multicolumn{1}{|l}{ \textit{Human presence}:} & \wbox &\multicolumn{6}{l|}{} \\
    \hline

    \rowcolor{gray!25}
    \multicolumn{8}{|l|}{\textbf{Datasets (RC1)}} \\

    \multicolumn{1}{|l}{\textit{Name:}} &
    \multicolumn{7}{l|}{\colorbox{lightgray}{Dataset name and version}} \\

    \multicolumn{1}{|l}{\textit{Record link*:}} &
    \multicolumn{7}{l|}{\colorbox{lightgray}{Persistent URL or DOI}} \\

    \multicolumn{1}{|l}{\textit{License:}} &
    \multicolumn{7}{l|}{\parbox{0.45\columnwidth}{\colorbox{lightgray}{Dataset license and usage restrictions}}} \\
    \hline

    \rowcolor{gray!25}
    \multicolumn{8}{|l|}{\textbf{Proposed method (RC2, RC3)}} \\

    \multicolumn{1}{|l}{\textit{Record link*:}} &
    \multicolumn{7}{l|}{\colorbox{lightgray}{Persistent URL or DOI}} \\

    \multicolumn{1}{|l}{\textit{Code link:}} &
    \multicolumn{7}{l|}{\colorbox{lightgray}{Repository URL and commit/tag}} \\

    \multicolumn{1}{|l}{\textit{Model card:}} &
    \multicolumn{7}{l|}{\wbox} \\

    \multicolumn{1}{|l}{\textit{License:}} &
    \multicolumn{7}{l|}{\colorbox{lightgray}{Model license and usage restrictions}} \\
    \hline

    \rowcolor{gray!25}
    \multicolumn{8}{|l|}{\textbf{Experimental setup (RC4)}} \\

    \multicolumn{1}{|l}{\textit{Data splits:}} &
    \multicolumn{7}{l|}{\colorbox{lightgray}{Train/validation/testing partitions}} \\

    \multicolumn{1}{|l}{\textit{Hyperparameters:}} &
    \multicolumn{7}{l|}{\colorbox{lightgray}{Learning rate, epochs, batch size}} \\

    \multicolumn{1}{|l}{\textit{Preprocessing:}} &
    \multicolumn{7}{l|}{\colorbox{lightgray}{Input size, resize and normalization}} \\
    \hline

    \rowcolor{gray!25}
    \multicolumn{8}{|l|}{\textbf{Performance measures (RC5)}} \\

    \multicolumn{1}{|l}{\textit{Metrics:}} &
    \multicolumn{7}{l|}{\colorbox{lightgray}{Evaluation metrics used}} \\

    \multicolumn{1}{|l}{\textit{Definition:}} &
    \multicolumn{7}{l|}{\colorbox{lightgray}{Metric formulation or reference}} \\

    \multicolumn{1}{|l}{\textit{Limitations:}} &
    \multicolumn{7}{l|}{\colorbox{lightgray}{Known assumptions and failure cases}} \\
    \hline

    \rowcolor{gray!25}
    \multicolumn{8}{|l|}{\textbf{Validation}} \\
    \multicolumn{1}{|l}{\textit{Generalisation:}} &
    \multicolumn{7}{l|}{\colorbox{lightgray}{Tests on unseen/novel objects, indoor/outdoor, clutter}} \\
    \multicolumn{1}{|l}{\textit{Robustness:}} &
    \multicolumn{7}{l|}{\colorbox{lightgray}{Tests varying occlusions, sensors artifacts, adversarial perturbations}} \\
    \multicolumn{1}{|l}{\textit{Robot deployment:}} & \multicolumn{7}{l|}{\wbox} \\
    \multicolumn{1}{|l}{\textit{Robot platform:}} &
    \multicolumn{7}{l|}{\colorbox{lightgray}{Robot model used for validation}} \\

    \multicolumn{1}{|l}{\textit{Robot end-effector:}} &
    \multicolumn{7}{l|}{\colorbox{lightgray}{Gripper or manipulator description}} \\

    \multicolumn{1}{|l}{\textit{Robot setup:}} &
    \multicolumn{7}{l|}{\colorbox{lightgray}{Simulation, laboratory or in-the-wild; robot task; success criteria}} \\

    \multicolumn{1}{|l}{\textit{Safety:}} &
    \multicolumn{7}{l|}{\colorbox{lightgray}{Standards/tests to assess the human safety based on model predictions}} \\
    \hline

    \multicolumn{8}{|l|}{
    \parbox{0.98\columnwidth}{
    \scriptsize
    \textit{Legend}: FUNC:~functional classification; FUNS:~functional segmentation;
    HPE:~hand pose estimation; HIS:~hand interaction synthesis;
    RC:~reproducibility challenge; '-': information not available.\\
    \textbf{Notes}: *Datasets, pre-trained weights and source code should be archived in repositories supporting persistent identifiers (e.g., DOI) to ensure long-term accessibility and reproducibility.
    }} \\
    \hline
    \end{tabular}
    \label{tab:affordance_sheet}
\end{table}

The second section describes the datasets (RC1) used by the proposed solution to detail their characteristics and share the link to the data and the license informing about data permissions. Future benchmarks should release a detailed description on how to use and visualize data so that researchers can get acquainted with the format, and should evaluate models under different conditions, such as generalization across unseen object instances, novel object categories, camera viewpoints, lighting conditions, clutter, partial occlusions, and different sensing modalities. Benchmarks for different tasks, such as COCO for object detection and instance segmentation~\cite{lin2014microsoft}, release only the training and validation sets, while keeping a private testing set not to bias the architecture designer. On the contrary, benchmarks for affordance prediction, such as UMD and IIT-AFF, release both the training and testing data, posing the risk of modifying the methods to improve performance scores on the testing set rather than formulating contributions that advance the field.

The third section highlights the model characteristics (RC2, RC3).
Providing model cards~\cite{mitchell2019model}, along with its implementation and trained weights, helps detail the description of models, supporting other researchers to build upon. When not available, we encourage the re-implementation and retraining of the models as a contribution for the community. Previous works~\cite{apicella2024segmenting, apicella2023affordance} re-implemented, retrained, and released models for affordance detection and segmentation due to the lack of publicly available models. We also recommend providing a link to the trained model's weights and a license detailing the allowed uses\footnote{Publicly-available software without a license is automatically protected by copyright. Other researchers cannot use the method implementation to reproduce results.
}. 

The fourth section provides the details of the experimental setup to train and evaluate methods under the same conditions (RC4), including pre-processing and post-processing information such as data splits, resize procedures, data normalisation, and hyper-parameters choice. The lack of these details can result in models with significantly different weights and in unfair comparisons.

The fifth section focuses on the criteria to validate and compare methods (RC5). Providing a stand-alone toolkit implementing the performance measures ensures the replicability of the results across different works. For affordance prediction, we recommend evaluating the performance of models using more than one measure to provide a more comprehensive analysis while identifying different aspects and limitations of the models. For example, in affordance segmentation, precision focuses on how many of the predicted pixels have the correct class and recall emphasizes how many of the annotated pixels are correctly predicted. Therefore, computing more than one score (and avoiding using a single score aggregating multiple performance measures) reduces the risk of drawing misleading conclusions that are based only on partial results.

The last section describes the potential validation of the method in terms of generalisation, robustness, safety, and robotic deployment. The Affordance Sheet documents the assessment of the model generalization across objects and environments unseen during training, robustness to realistic sensing disturbances such as occlusions and illumination changes, and safety through standardized tests and protocols for interaction with humans. In previous works, few of the methods were validated using a real robotic platform~\cite{nguyen2017object,do2018affordancenet,yin2022object}. When a robot experiment can be performed, we recommend reporting the characteristics of the setup, the robotic hand specifics, and the description of the experiment in terms of object, conditions, and physics engine in case of simulated robot. This transparent reporting allows researchers to assess methods using a common platform. All these aspects promote more transparent assessment of multimodal affordance models in real-world, human-centred scenarios.

\section{Comparison of Previous Methods}

We provide examples of Affordance Sheets (empty fields are not reported): EgoTopo~\cite{nagarajan2020ego} for functional classification in Table~\ref{tab:affsheet_egotopo}, ACANet~\cite{apicella2023affordance} for affordance segmentation in Table~\ref{tab:affsheet_acanet}, Multi-FinGAN~\cite{lundell2021multi} and HUG~\cite{wu2026human} for hand pose estimation in Table~\ref{tab:affsheet_multifingan} and Table~\ref{tab:affsheet_hug}, respectively. These examples highlight substantial differences in the  practices adopted by the methods, revealing reproducibility gaps difficult to identify from the original publications alone. 

\begin{table}[t!]
    \centering
    \setlength\tabcolsep{2pt}
    \scriptsize
    \caption{Affordance Sheet for EgoTopo~\cite{nagarajan2020ego}.}
    \begin{tabular}{|c c c c c c c c|}
    \hline
    \rowcolor{gray!50}
    \multicolumn{8}{|c|}{\textbf{EgoTopo (15/07/2026)}} \\
    \hline
    \rowcolor{gray!25}
    \multicolumn{8}{|l|}{\textbf{Formulation}} \\
    \multicolumn{1}{|l}{\textit{Affordance task:}} 
     & FUNC & FUNS & HPE & HIS & & & \\
     & \bbox & \wbox & \wbox & \wbox & & & \\
    \multicolumn{1}{|l}{\textit{Modality:}} & Image & Video & Language & Point cloud & Depth & Audio & Tactile \\
     & \bbox & \wbox & \wbox & \wbox & \wbox & \wbox & \wbox \\
    \multicolumn{1}{|l}{ \textit{Human presence}:} & \wbox &\multicolumn{6}{l|}{} \\
    \hline
    \rowcolor{gray!25}
    \multicolumn{8}{|l|}{\textbf{Datasets (RC1)}} \\
    \multicolumn{1}{|l}{\textit{Name:}} & \multicolumn{7}{l|}{EPIC-Kitchens} \\ 
    \multicolumn{1}{|l}{\textit{Record link:}} & \multicolumn{7}{l|}{\parbox{0.7\columnwidth}{\url{https://data.bris.ac.uk/data/dataset/2g1n6qdydwa9u22shpxqzp0t8m}}} \\
    \multicolumn{1}{|l}{\textit{License:}} & \multicolumn{7}{l|}{\parbox{0.4\columnwidth}{\strut CC-BY-NC 4.0 \strut}}\\  
    \hline
    \rowcolor{gray!25}
    \multicolumn{8}{|l|}{\textbf{Proposed method (RC2, RC3)}} \\
    \multicolumn{1}{|l}{\textit{Record link:}} & \multicolumn{7}{l|}{\parbox{0.57\columnwidth}{\url{https://dl.fbaipublicfiles.com/ego-topo/anticipation/pretrained.zip}}} \\ 
    \multicolumn{1}{|l}{\textit{Code link:}} & \multicolumn{7}{l|}{\url{https://github.com/facebookresearch/ego-topo}} \\
    \multicolumn{1}{|l}{ \textit{Model card}:} & \multicolumn{7}{l|}{\wbox} \\
    \multicolumn{1}{|l}{\textit{License:}} & \multicolumn{7}{l|}{CC-BY-NC 4.0} \\  
    \hline
    \rowcolor{gray!25}
    \multicolumn{8}{|l|}{\textbf{Experimental setup (RC4)}} \\
    \multicolumn{8}{|l|}{\textit{Data splits:}} \\
    &  \multicolumn{7}{l|}{ 
    \begin{tabular}{l c}
    \toprule
    Set & Images \\
    \midrule
    Training & - \\
    Validation & - \\
    Testing & 1,155 \\
    \bottomrule
    \end{tabular}}
    \\
    \multicolumn{8}{|l|}{\textit{Hyperparameters:}} \\
    & \multicolumn{7}{l|}{
    \begin{tabular}{l c}
    \toprule
    Name & Value \\
    \midrule
    epochs & 20 \\
    batch size & 256 \\
    learning rate & 0.0001 \\
    schedule & 0.1x after 15 epochs \\
    patience & - \\ 
    optimizer & Adam \\
    momentum & -  \\ 
    weight decay & 0.000001 \\ 
    resize & - \\
    flip & - \\
    \bottomrule
    \end{tabular}
    }\\
    \multicolumn{1}{|l}{\textit{Resize procedure:}} & \multicolumn{7}{l|}{-}\\  
    \hline
    \rowcolor{gray!25}
    \multicolumn{8}{|l|}{\textbf{Performance measures (RC5)}} \\
    \multicolumn{1}{|l}{\textit{Description:}} & \multicolumn{7}{l|}{\parbox{0.7\columnwidth}{\strut Mean  average precision (mAP) over all afforded interactions. \strut}}\\  
    \hline
    \rowcolor{gray!25}
    \multicolumn{8}{|l|}{\textbf{Validation}} \\
    \multicolumn{1}{|l}{\textit{Robot deployment:}} & \multicolumn{7}{l|}{\wbox} \\
    \hline
    \end{tabular}
    \label{tab:affsheet_egotopo}
\end{table}
\begin{table}[t!]
    \centering
    \setlength\tabcolsep{2pt}
    \scriptsize
    \caption{Affordance sheet for ACANet~\cite{apicella2023affordance}.}
    \begin{tabular}{|c c c c c c c c|}
    \hline
    \rowcolor{gray!50}
    \multicolumn{8}{|c|}{\textbf{ACANet (15/07/2026)}} \\
    \hline
    \rowcolor{gray!25}
    \multicolumn{8}{|l|}{\textbf{Formulation}} \\
    \multicolumn{1}{|l}{\textit{Affordance task:}} 
     & FUNC & FUNS & HPE & HIS & & & \\
     & \wbox & \bbox & \wbox & \wbox & & & \\
     \multicolumn{1}{|l}{\textit{Modality:}} & Image & Video & Language & Point cloud & Depth & Audio & Tactile \\
     & \bbox & \wbox & \wbox & \wbox & \wbox & \wbox & \wbox \\
    \multicolumn{1}{|l}{ \textit{Human presence}:} & \bbox &\multicolumn{6}{l|}{} \\
    \hline
    \rowcolor{gray!25}
    \multicolumn{8}{|l|}{\textbf{Datasets (RC1)}} \\
    \multicolumn{1}{|l}{\textit{Name:}} & \multicolumn{7}{l|}{CHOC-AFF} \\ 
    \multicolumn{1}{|l}{\textit{Record link:}} & \multicolumn{7}{l|}{\parbox{0.6\columnwidth}{\url{https://doi.org/10.5281/zenodo.5085800}}} \\
    \multicolumn{1}{|l}{\textit{License:}} & \multicolumn{7}{l|}{\parbox{0.4\columnwidth}{\strut CC BY 4.0 \strut}}\\  
    \hline
    \rowcolor{gray!25}
    \multicolumn{8}{|l|}{\textbf{Proposed method (RC2, RC3)}} \\
    \multicolumn{1}{|l}{\textit{Record link:}} & \multicolumn{7}{l|}{\parbox{0.5\columnwidth}{\url{https://doi.org/10.5281/zenodo.8364196}}} \\ 
    \multicolumn{1}{|l}{\textit{Code link:}} & \multicolumn{7}{l|}{\url{https://github.com/apicis/aff-seg/}} \\
    \multicolumn{1}{|l}{ \textit{Model card}:} & \multicolumn{7}{l|}{\bbox} \\
    \multicolumn{1}{|l}{\textit{License:}} & \multicolumn{7}{l|}{CC BY-NC-SA 4.0} \\  
    \hline
    \rowcolor{gray!25}
    \multicolumn{8}{|l|}{\textbf{Experimental setup (RC4)}} \\
    \multicolumn{8}{|l|}{\textit{Data splits:}} \\
    &  \multicolumn{7}{l|}{ 
     \begin{tabular}{l c}
    \toprule
    Set & Images \\
    \midrule
    Training & 89,856 \\
    Validation & 17,280 \\
    Testing 1 & 13,824 \\
    Testing 2 & 17,280 \\
    \bottomrule
    \end{tabular}}
    \\
    \multicolumn{8}{|l|}{\textit{Hyperparameters:}} \\
    & \multicolumn{7}{l|}{
    \begin{tabular}{l c}
    \toprule
    Name & Value \\
    \midrule
    batch size & 2 \\
    learning rate & 0.001 \\
    schedule & 0.5x \\
    patience & 3 \\ 
    optimizer & SGD \\
    momentum & 0.9  \\ 
    weight decay & 0.0001 \\ 
    resize & [1, 1.5] \\
    flip & 0.5 \\
    \bottomrule
    \end{tabular}
    }\\
    \multicolumn{1}{|l}{\textit{Resize procedure:}} & \multicolumn{7}{l|}{center crop  $480 \times 480$}\\  
    \hline
    \rowcolor{gray!25}
    \multicolumn{8}{|l|}{\textbf{Performance measures (RC5)}} \\
    \multicolumn{1}{|l}{\textit{Description:}} & \multicolumn{7}{l|}{\parbox{0.75\columnwidth}{\strut Per-class Jaccard index measures the overlap between predicted and annotated segmentation masks, and quantifies how much they are similar in size \strut}}\\  
    \multicolumn{1}{|l}{\textit{Formulation:}} & \multicolumn{7}{l|}{\parbox{0.6\linewidth}{$\frac{\sum_{n=1}^{N} \sum_{\boldsymbol{y} \in I_n} TP^{\boldsymbol{y}}_n}{ \sum_{n=1}^{N} \sum_{\boldsymbol{y} \in I_n} TP^{\boldsymbol{y}}_n + FP^{\boldsymbol{y}}_n + FN^{\boldsymbol{y}}_n}$}}\\  
    \hline
    \rowcolor{gray!25}
    \multicolumn{8}{|l|}{\textbf{Validation}} \\
    \multicolumn{1}{|l}{\textit{Generalisation:}} & \multicolumn{7}{l|}{\parbox{0.75\columnwidth}{\strut Model trained on mixed-reality data and tested on 300 real images with different object instances \strut}} \\
    \multicolumn{1}{|l}{\textit{Robot deployment:}} & \multicolumn{7}{l|}{\wbox} \\
    \hline
    \end{tabular}
    \label{tab:affsheet_acanet}
\end{table}
\begin{table}[t!]
    \centering
    \setlength\tabcolsep{2pt}
    \scriptsize
    \caption{Affordance Sheet for Multi-FinGAN~\cite{lundell2021multi}.}
    \begin{tabular}{|c c c c c c c c |}
    \hline
    \rowcolor{gray!50}
    \multicolumn{8}{|c|}{\textbf{Multi-FinGAN (15/07/2026)}} \\
    \hline
    \rowcolor{gray!25}
    \multicolumn{8}{|l|}{\textbf{Formulation}} \\
    \multicolumn{1}{|l}{\textit{Affordance task:}} 
     & FUNC & FUNS & HPE & HIS & & & \\
    & \wbox & \wbox & \bbox & \wbox & & &  \\
    \multicolumn{1}{|l}{\textit{Modality:}} & Image & Video & Language & Point cloud & Depth & Audio & Tactile \\
     & \bbox & \wbox & \wbox & \wbox & \wbox & \wbox & \wbox \\
    \multicolumn{1}{|l}{ \textit{Human presence}:} & \wbox &\multicolumn{6}{l|}{} \\
    \hline
    \rowcolor{gray!25}
    \multicolumn{8}{|l|}{\textbf{Datasets (RC1)}} \\
    \multicolumn{1}{|l}{\textit{Name:}} & \multicolumn{7}{l|}{-} \\ 
    \multicolumn{1}{|l}{\textit{Record link:}} & \multicolumn{7}{l|}{\parbox{0.7\columnwidth}{\url{https://github.com/aalto-intelligent-robotics/Multi-FinGAN/blob/main/data/download_train_data.sh}}} \\
    \multicolumn{1}{|l}{\textit{License:}} & \multicolumn{7}{l|}{\parbox{0.4\columnwidth}{\strut - \strut}}\\  
    \hline
    \rowcolor{gray!25}
    \multicolumn{8}{|l|}{\textbf{Proposed method (RC2, RC3)}} \\
    \multicolumn{1}{|l}{\textit{Record link:}} & \multicolumn{7}{l|}{\parbox{0.7\columnwidth}{\url{https://drive.google.com/file/d/19462M8s3tEXe_1_riHuvQegLxzdX-kl2/view}}} \\ 
    \multicolumn{1}{|l}{\textit{Code link:}} & \multicolumn{7}{l|}{\url{https://github.com/aalto-intelligent-robotics/Multi-FinGAN}} \\
    \multicolumn{1}{|l}{ \textit{Model card}:} & \multicolumn{7}{l|}{\wbox} \\
    \multicolumn{1}{|l}{\textit{License:}} & \multicolumn{7}{l|}{MIT} \\  
    \hline
    \rowcolor{gray!25}
    \multicolumn{8}{|l|}{\textbf{Experimental setup (RC4)}} \\
    \multicolumn{8}{|l|}{\textit{Data splits:}} \\
    &  \multicolumn{7}{l|}{ 
    \begin{tabular}{l c}
    \toprule
    Set & Images \\
    \midrule
    Training & 3000 \\
    Validation & - \\
    Testing & - \\
    \bottomrule
    \end{tabular}}
    \\
    \multicolumn{8}{|l|}{\textit{Hyperparameters:}} \\
    & \multicolumn{7}{l|}{
    \begin{tabular}{l c}
    \toprule
    Name & Value \\
    \midrule
    batch size & 100 \\
    learning rate & 0.0001 \\
    schedule & linear after 400 epochs \\
    patience & - \\ 
    optimizer & Adam \\
    momentum & default  \\ 
    weight decay & default \\ 
    resize & - \\
    flip & - \\
    \bottomrule
    \end{tabular}
    }\\
    \multicolumn{1}{|l}{\textit{Resize procedure:}} & \multicolumn{7}{l|}{Object centric crops resized to $256 \times 256$}\\  
    \hline
    \rowcolor{gray!25}
    \multicolumn{8}{|l|}{\textbf{Performance measures (RC5)}} \\
    \multicolumn{1}{|l}{\textit{Description:}} & \multicolumn{7}{l|}{\parbox{0.6\columnwidth}{\strut Interpenetration: amount of voxels in common between object and hand. \strut}}\\  
    \hline
    \rowcolor{gray!25}
    \multicolumn{8}{|l|}{\textbf{Validation}} \\
    \multicolumn{1}{|l}{\textit{Generalisation:}} & \multicolumn{7}{l|}{\parbox{0.7\columnwidth}{\strut Model trained on synthetic data and tested on real images. \strut}} \\
    \multicolumn{1}{|l}{\textit{Robot deployment:}} & \multicolumn{7}{l|}{\bbox} \\
    \multicolumn{1}{|l}{\textit{Robot platform:}} & \multicolumn{7}{l|}{\parbox{0.35\columnwidth}{Franka Emika Panda}}\\
    \multicolumn{1}{|l}{\textit{Robot end-effector:}} & \multicolumn{7}{l|}{\parbox{0.35\columnwidth}{Barrett hand}}\\
    \multicolumn{1}{|l}{\textit{Robot setup:}} & \multicolumn{7}{l|}{\parbox{0.78\columnwidth}{Intel RealSense D435 camera looking at the scene at 45 degree viewpoint. The model generates 20 grasps per object and then intersection and quality metric of each grasp are computed. The first physically reachable grasp with lowest intersection and highest quality metric is executed on the real robot. The robot needs to grasp the object and, without dropping it, move to the start position and rotate the hand ±90° around the last joint (success). If the object was dropped during the manipulation, the grasp is considered unsuccessful.}}\\
    \hline
    \end{tabular}
    \label{tab:affsheet_multifingan}
\end{table}
\begin{table}[t!]
    \centering
    \setlength\tabcolsep{2pt}
    \scriptsize
    \caption{Affordance Sheet for HUG~\cite{wu2026human}.}
    \begin{tabular}{|c c c c c c c c|}
    \hline
    \rowcolor{gray!50}
    \multicolumn{8}{|c|}{\textbf{HUG (15/07/2026)}} \\
    \hline
    \rowcolor{gray!25}
    \multicolumn{8}{|l|}{\textbf{Formulation}} \\
    \multicolumn{1}{|l}{\textit{Affordance task:}} 
     & FUNC & FUNS & HPE & HIS & & & \\
    & \wbox & \wbox & \bbox & \wbox & & & \\
    \multicolumn{1}{|l}{\textit{Modality:}} & Image & Video & Language & Point cloud & Depth & Audio & Tactile \\
     & \bbox & \wbox & \wbox & \wbox & \bbox & \wbox & \wbox \\
    \multicolumn{1}{|l}{ \textit{Human presence}:} & \wbox &\multicolumn{6}{l|}{} \\
    \hline
    \rowcolor{gray!25}
    \multicolumn{8}{|l|}{\textbf{Datasets (RC1)}} \\
    \multicolumn{1}{|l}{\textit{Name:}} & \multicolumn{7}{l|}{1M-HUGS} \\ 
    \multicolumn{1}{|l}{\textit{Record link:}} & \multicolumn{7}{l|}{\parbox{0.7\columnwidth}{-}} \\
    \multicolumn{1}{|l}{\textit{License:}} & \multicolumn{7}{l|}{\parbox{0.4\columnwidth}{\strut - \strut}}\\  
    \hline
    \rowcolor{gray!25}
    \multicolumn{8}{|l|}{\textbf{Proposed method (RC2, RC3)}} \\
    \multicolumn{1}{|l}{\textit{Record link:}} & \multicolumn{7}{l|}{\parbox{0.7\columnwidth}{\url{https://huggingface.co/kevinywu/hug}}} \\ 
    \multicolumn{1}{|l}{\textit{Code link:}} & \multicolumn{7}{l|}{\url{https://github.com/KevinyWu/hug}} \\
    \multicolumn{1}{|l}{ \textit{Model card}:} & \multicolumn{7}{l|}{\wbox} \\
    \multicolumn{1}{|l}{\textit{License:}} & \multicolumn{7}{l|}{MIT} \\  
    \hline
    \rowcolor{gray!25}
    \multicolumn{8}{|l|}{\textbf{Experimental setup (RC4)}} \\
    \multicolumn{8}{|l|}{\textit{Data splits:}} \\
    &  \multicolumn{7}{l|}{ 
    \begin{tabular}{l c}
    \toprule
    Set & Images \\
    \midrule
    Training & 1M \\
    Validation & - \\
    Testing & - \\
    \bottomrule
    \end{tabular}}
    \\
    \multicolumn{8}{|l|}{\textit{Hyper-parameters:}} \\
    & \multicolumn{7}{l|}{
    \begin{tabular}{l c}
    \toprule
    Name & Value \\
    \midrule
    batch size & 128 \\
    learning rate & 0.0001 \\
    schedule & linear for 5K steps \\
    patience & - \\ 
    optimizer & AdamW \\
    momentum & default  \\ 
    weight decay & 0.001 \\ 
    resize & - \\
    flip & - \\
    \bottomrule
    \end{tabular}
    }\\
    \multicolumn{1}{|l}{\textit{Resize procedure:}} & \multicolumn{7}{l|}{Crops resized to $224 \times 224$}\\  
    \hline
    \rowcolor{gray!25}
    \multicolumn{8}{|l|}{\textbf{Performance measures (RC5)}} \\
    \multicolumn{1}{|l}{\textit{Description:}} & \multicolumn{7}{l|}{\parbox{0.6\columnwidth}{\strut Fingertip contact error (mm): how close the thumb and the closest supporting finger come to the object surface. \strut}}\\  
    \multicolumn{1}{|l}{\textit{Formulation:}} & \multicolumn{7}{l|}{\parbox{0.6\linewidth}{
    $ FC = \frac{1}{2} (|d_{thumb}| + \min_{f \in \mathcal{F}} (d_f))$, where $d$ is signed distance and $\mathcal{F}$ non-thumb fingers set }}\\     
    \hline
    \rowcolor{gray!25}
    \multicolumn{8}{|l|}{\textbf{Validation}} \\
    \multicolumn{1}{|l}{\textit{Generalisation:}} &
    \multicolumn{7}{l|}{\parbox{0.78\columnwidth}{HUG is tested in real and simulated environments with 90 unseen objects belonging to 5 geometric categories (cylindrical, spheroidal, prismatic, appendaged, amorphous) and 3 size bins (small, medium, large).}} \\
    \multicolumn{1}{|l}{\textit{Robot deployment:}} & \multicolumn{7}{l|}{\bbox} \\
    \multicolumn{1}{|l}{\textit{Robot platform:}} & \multicolumn{7}{l|}{\parbox{0.35\columnwidth}{7-DoF xArm}}\\
    \multicolumn{1}{|l}{\textit{Robot end-effector:}} & \multicolumn{7}{l|}{\parbox{0.35\columnwidth}{6-DoF Ability hand}}\\
    \multicolumn{1}{|l}{\textit{Robot setup:}} & \multicolumn{7}{l|}{\parbox{0.7\columnwidth}{An external ZED camera is the input to the model that controls the robot. The output of the model (MANO grasp) is mapped to a target robot hand at deployment. Each robot hand’s fingertips is mapped to MANO’s using a single fixed offset, estimated from simulation.}}\\
    \hline
    \end{tabular}
    \label{tab:affsheet_hug}
\end{table}

EgoTopo, MultiFin-Gan, and ACANet provide access to the training data while the details about metadata such as dataset versions, licenses, and persistent identifiers is inconsistent, complicating long-term reproducibility and dataset reuse. The Affordance Sheets also reveal notable differences in the reporting of experimental protocols. ACANet provides the most complete description, including training, validation and multiple test splits, extensive hyper-parameters, and image preprocessing details. In contrast, EgoTopo omits the training and validation split sizes and several preprocessing parameters, while Multi-FinGAN does not report validation or testing partitions and leaves several training hyper-parameters unspecified. Such inconsistencies make it difficult to reproduce published results or determine whether observed performance differences arise from methodological innovations or from variations in the experimental setup.
Similarly, the reported evaluation protocols vary substantially. ACANet and HUG provide a formal definition of the Jaccard Index and Fingertip contact error, whereas EgoTopo and Multi-FinGAN only describe their evaluation metrics without specifying their mathematical formulation. No method described the limitations of the chosen performance measures. Explicitly documenting the assumptions and limitations of performance measures is important because different metrics capture different performance aspects, favouring different model behaviours.
Neither EgoTopo nor ACANet describe any robotic deployment, leaving open the question of how their predictions transfer to real manipulation scenarios. In contrast, Multi-FinGAN and HUG include a real-world evaluation by specifying the robot platform, end-effector, sensing setup, execution protocol, and success criterion. HUG's generalisation to different objects shape is assessed using 90 object categories unseen during training with different sizes and geometries. The generalisation of ACANet and Multi-FinGAN is tested on real images, different from those in the training data (synthetic or mixed-reality images). None of the methods assessed robustness or human safety.

\section{Conclusion}

In this work, we identified the factors that undermine fair benchmarking, reproducible evaluation, and deployment-oriented assessment of multimodal models for affordance prediction: the diversity of problem formulations, the limited availability of implementations, and the insufficient reporting of experimental setups. 
To promote transparency, we introduced the Affordance Sheet, a tool designed to systematically document models, datasets, and validation protocols, focusing on generalisation to diverse objects and environments, robustness to occlusions, and scenarios where the agent interacts with humans (human safety).
Affordance Sheets aim to facilitate comparable research outcomes beyond quantitative performance and expose missing information that directly affects reproducibility and fair benchmarking.
Affordance Sheet represents the first example of reporting practice for multimodal affordance prediction models, and supports a more informed design of the next-generation of learning-based robotic systems operating reliably in diverse real-world environments. The tool can also be generalised to  tasks different from affordance prediction and that would require similar transparent documentation.  

\small

\bibliographystyle{splncs04}
\bibliography{ax_short_strings,refs}

\end{document}